\documentclass{article}

\usepackage{microtype}
\usepackage{graphicx}
\usepackage{subfigure}
\usepackage{booktabs} % for professional tables
\usepackage{caption}
\usepackage{subcaption}
\usepackage{multirow}
\usepackage{multicol}

\usepackage{hyperref}

\usepackage[accepted]{icml2025}

\usepackage{amsmath}
\usepackage{amssymb}
\usepackage{mathtools}
\usepackage{amsthm}

\usepackage[capitalize,noabbrev]{cleveref}

\theoremstyle{plain}

\theoremstyle{definition}

\theoremstyle{remark}

\usepackage[textsize=tiny]{todonotes}

\icmltitlerunning{Diffusion Decoding for Peptide De Novo Sequencing}

\begin{document}

\twocolumn[
\icmltitle{Diffusion Decoding for Peptide De Novo Sequencing}




\begin{icmlauthorlist}
\icmlauthor{Chi-en Amy Tai}{yyy}
\icmlauthor{Alexander Wong}{yyy}
\end{icmlauthorlist}

\icmlaffiliation{yyy}{University of Waterloo, Waterloo, Canada}

\icmlcorrespondingauthor{Chi-en Amy Tai} {amy.tai@uwaterloo.ca}

\icmlkeywords{Machine Learning, ICML}

\vskip 0.3in
]



\printAffiliationsAndNotice{} % otherwise use the standard text.

\begin{abstract}
Peptide de novo sequencing is a method used to reconstruct amino acid sequences from tandem mass spectrometry data without relying on existing protein sequence databases. Traditional deep learning approaches, such as Casanovo, mainly utilize autoregressive decoders and predict amino acids sequentially. Subsequently, they encounter cascading errors and fail to leverage high-confidence regions effectively. To address these issues, this paper investigates using diffusion decoders adapted for the discrete data domain. These decoders provide a different approach, allowing sequence generation to start from any peptide segment, thereby enhancing prediction accuracy. We experiment with three different diffusion decoder designs, knapsack beam search, and various loss functions. We find knapsack beam search did not improve performance metrics and simply replacing the transformer decoder with a diffusion decoder lowered performance. Although peptide precision and recall were still 0, the best diffusion decoder design with the DINOISER loss function obtained a statistically significant improvement in amino acid recall by 0.373 compared to the baseline autoregressive decoder-based Casanovo model. These findings highlight the potential of diffusion decoders to not only enhance model sensitivity but also drive significant advancements in peptide de novo sequencing.
\end{abstract}

\section{Introduction}
\begin{figure}
    \centering
    \includegraphics[width=.9\linewidth]{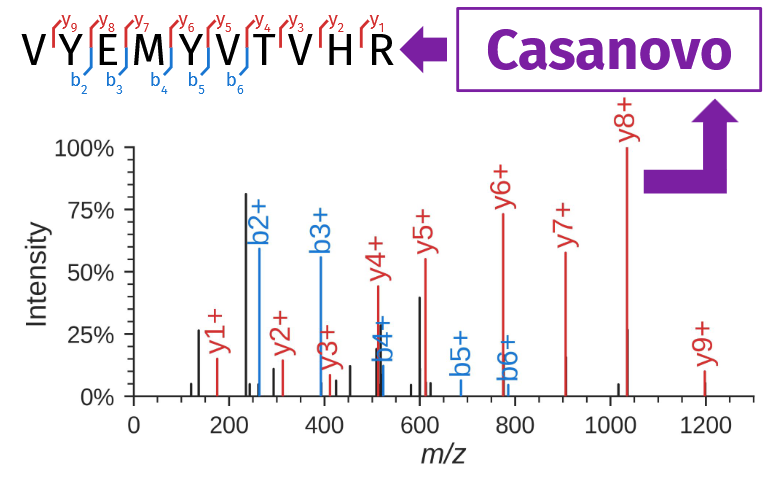}
    \caption{Overview of the peptide de novo sequencing process, illustrating the flow from inputting an observed spectrum to generating the corresponding peptide sequence, copied from Casanovo~\cite{casanovo-yilmaz2022novo}.}
  \label{fig:peptide-sequencing-flow}
\end{figure}

% Provide background information on the topic. Clearly state the problem you are addressing and the objectives of your project. Explain the significance of the project in the context of deep learning.
Peptide de novo sequencing is the task of reconstructing the amino acid sequence of peptides directly from tandem mass spectrometry (MS/MS) data, without relying on existing protein sequence databases~\cite{deepnovo-tran2017novo}. It is particularly valuable for identifying novel peptides or sequences from organisms with unsequenced genomes~\cite{graphnovo-mao2023mitigating}. Obtaining high accuracy is challenging due to incomplete fragmentation of precursor peptides and noise present in MS/MS spectra (seen as black peaks in Figure~\ref{fig:peptide-sequencing-flow}~\cite{deepnovo-tran2017novo}. Deep learning offers innovative solutions to these challenges by efficiently modeling intricate relationships within spectral data~\cite{deepnovo-tran2017novo}. Consequently, recent breakthroughs in deep learning have surpassed earlier methods, which relied on heuristic search and dynamic programming, delivering enhanced accuracy and efficiency~\cite{casanovo-yilmaz2022novo}.

\begin{figure}
    \centering
    \includegraphics[width=.9\linewidth]{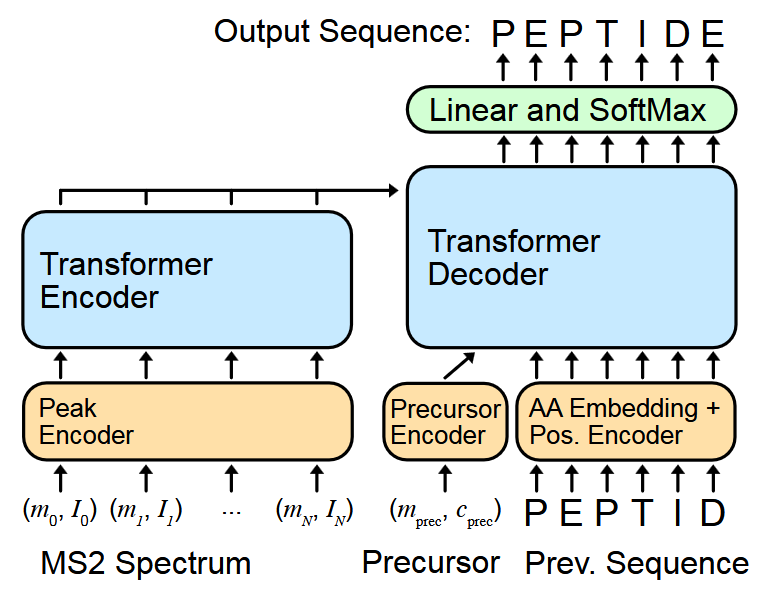}
    \caption{Sample deep learning model architecture, copied from Casanovo~\cite{casanovo-yilmaz2022novo}.}
  \label{fig:peptide-sequencing-casanovo}
\end{figure}

However, existing deep learning models for peptide sequencing predominantly use autoregressive decoders inspired by natural language processing frameworks~\cite{casanovo-yilmaz2022novo,graphnovo-mao2023mitigating} with an example seen in Figure~\ref{fig:peptide-sequencing-casanovo}. Autoregressive decoders sequentially generate sequences of amino acids where each prediction is dependent on its predecessor~\cite{casanovo-yilmaz2022novo}. In the context of peptide de novo sequencing, this behaviour is problematic as early errors could cascade through the entire sequence and it does not take advantage of performance gains from decoding high confidence parts first (often the middle part of the peptide)~\cite{deepnovo-tran2017novo}. Using a diffusion decoder, especially one adapted for the discrete data domain, is a promising alternative as diffusion decoders can start decoding from any part of the peptide sequence, allowing them to capitalize on high confidence parts to enhance overall prediction accuracy~\cite{lou2023discrete}. 

This paper addresses the challenge of using diffusion decoding for peptide de novo sequencing and explores how to better integrate diffusion in the deep learning model architecture. The objectives of the project include incorporating different diffusion decoder designs into peptide sequencing workflows to replace traditional autoregressive decoders, exploring the compatibility of diffusion decoding with knapsack beam search, and evaluating the effectiveness of different loss functions compared to traditional cross-entropy. This work highlights how deep learning is vital in overcoming challenges in noisy MS/MS data and how diffusion decoding is a promising direction for more accurate peptide sequencing processes. 

\section{Related Work}
% Review relevant literature and existing approaches to the problem. Highlight gaps or limitations in current research that your project aims to address.

% Does the report provide a comprehensive review of related work with key references?
% Does it clearly identify gaps or limitations in current research?
% Does it effectively position the project within the context of existing literature?

\subsection{Deep Learning Models for Peptide De Novo Sequencing}
DeepNovo was the first deep neural network model proposed for peptide de novo sequencing~\cite{deepnovo-tran2017novo}. DeepNovo combined convolutional neural networks and long short-tappearingerm memory recurrent neural networks with local dynamic programming to predict peptide sequences from MS/MS data~\cite{deepnovo-tran2017novo}. Dynamic programming using the knapsack approach was applied to exclude amino acids whose masses exceeded the suffix mass~\cite{deepnovo-tran2017novo}. The final heuristic for exploring candidate sequences, however, was beam search, which evaluated a predetermined set of the most promising sequences at each iteration to determine the optimal prediction~\cite{deepnovo-tran2017novo}. Compared to state-of-the-art methods at the time, DeepNovo achieved higher accuracies at the amino acid and peptide level, but it was noted that the model performance was heavily dependent on the dataset~\cite{deepnovo-tran2017novo}. Moreover, DeepNovo had to strike a balance between low binning resolution, which reduced sequencing accuracy, and increased model complexity, which caused longer inference times~\cite{casanovo-yilmaz2022novo} as it needed to discretize the mass-to-charge (m/z) axis of the mass spectra during sequencing~\cite{casanovo-yilmaz2022novo}.

% Talk about PointNovo? => leave for now as space may be issue

Casanovo instead eliminated the need for discretization and used both the precursor mass and charge, along with individual spectrum peaks, as inputs to its model~\cite{casanovo-yilmaz2022novo}. Casanovo was designed using transformers with the self-attention mechanism to translate observed spectrum peaks into amino acid sequences and the model was trained using supervised learning where database searches supplied the ground truth sequences for prediction~\cite{casanovo-yilmaz2022novo}. Instead of relying on dynamic programming, a delta mass filter was applied to process the peptide sequences~\cite{casanovo-yilmaz2022novo}. Notably the knapsack dynamic programming algorithm was tested for post-processing but the simple m/z filter yielded better performance~\cite{casanovo-yilmaz2022novo}. Beam search was subsequently implemented in the publicly available code to obtain the optimal prediction~\cite{casanovo-yilmaz2022novo}. Based on experiments using a cross-species evaluation benchmark, Casanovo achieved better performance with lower inference time and complexity compared to DeepNovo~\cite{casanovo-yilmaz2022novo}. However, some Casanovo predictions could be eliminated by the delta mass filter, resulting in gaps in plausible predictions for significant portions of the spectra~\cite{casanovo-yilmaz2022novo}.

GraphNovo is a two-stage graph-based deep learning model with the goal to tackle the problem of missing fragmentation for peptide de novo sequencing~\cite{graphnovo-mao2023mitigating}. A graph is constructed from the spectral data and the first stage focuses on identifying an optimal path using the graph~\cite{graphnovo-mao2023mitigating}. This optimal path is mapped to a peptide sequence containing mass tags for missing fragmentation regions~\cite{graphnovo-mao2023mitigating}. In the second stage, these mass tags are processed and replaced with predicted amino acids to produce the final peptide sequence~\cite{graphnovo-mao2023mitigating}. Both stages use a transformer structure as the encoder and decoder and its design subsequently required high computational power; training needed four A100 graphic processing units (GPUs)~\cite{graphnovo-mao2023mitigating}. Since GraphNovo relied on transformer decoders, a limitation also exists for predicting long sequences in that early errors could cascade through the entire sequence. In addition, GraphNovo does not incorporate all the provided information such as isotope peaks which could provide performance gains. Similar to DeepNovo, GraphNovo also used the knapsack algorithm for filtering and the beam search technique for final peptide generation~\cite{graphnovo-mao2023mitigating}.

InstaNovo is a transformer model recently introduced for de novo sequencing and proposes an iterative refinement step using InstaNovo+, its diffusion-powered model~\cite{instanovo-eloff2025instanovo}. Given predicted sequences from InstaNovo, InstaNovo+ refines and improves the sequence by treating it as a corrupted sequence and iteratively removing noise~\cite{instanovo-eloff2025instanovo}. Key elements of InstaNovo are its use of knapsack beam search decoding and multi-scale sinusoidal embeddings to improve performance, but at high computational cost~\cite{instanovo-eloff2025instanovo}. In addition, InstaNovo+ is seen as a secondary step after InstaNovo is run and requires multiple rounds of refinement for prediction, making it computationally expensive and time-consuming with the performance of InstaNovo+ limited by the performance of InstaNovo~\cite{instanovo-eloff2025instanovo}. Furthermore, InstaNovo+ is trained using average KL-divergence loss of the model and they do not explore training with other loss functions which could be more appropriate for discrete diffusion modelling~\cite{instanovo-eloff2025instanovo}. 

\subsection{Discrete Diffusion Modelling Techniques}
Diffusion models are widely used in generative research because they simulate data generation via a stochastic process, transforming random noise into structured data through iterative denoising steps to produce high-quality outputs~\cite{ho2020denoising}. During training, the forward step adds controlled noise to structured data~\cite{ho2020denoising}. Then, to create coherent outputs, noise is removed step-by-step~\cite{ho2020denoising}. Initially, diffusion models were implemented for continuous data domains such as images and were subsequently less effective for discrete data domains like text. 

Diffusion-LM was introduced to tackle the challenge of text generation in the discrete domain by utilizing continuous diffusion models~\cite{li2022diffusion}. By transforming text sequences into continuous embeddings, Diffusion-LM incrementally refines a sequence of Gaussian noise vectors into meaningful word representations~\cite{li2022diffusion}. The model achieves this through a "rounding" method, where it selects the word with the highest likelihood and adjusts the embeddings to correspond to actual token embeddings at each step of the diffusion process~\cite{li2022diffusion}. However, this approach relies heavily on continuous diffusion processes and has significant computational demands during both training and inference due to the rounding step.

Score Entropy Discrete Diffusion models (SEDD) shifted away from converting discrete data into a continuous space for modelling and instead, operated entirely in the discrete token space using probability vectors and transition matrices. SEDD is thus controllable, allowing for prompting from arbitrary positions~\cite{lou2023discrete}. In text-based experiments, SEDD outperformed earlier discrete and continuous diffusion language models on both standard (left-to-right) generation and infilling tasks, surpassing strong autoregressive models such as GPT-2~\cite{lou2023discrete}. Notably, SEDD is computationally intensive and needs 8 A100 80 GB GPUs for training. 

DINOISER is a less computationally demanding method that also tries to tackle the problem of diffusion modelling for discrete data~\cite{ye2023dinoiser}. Unlike SEDD, DINOISER builds on Diffusion-LM and focuses on manipulating noises~\cite{ye2023dinoiser}. DINOISER uses noise scale clipping to avoid training on small noise scales for better generalization and was shown to achieve state-of-the-art results compared to other models like DiffusionLM for machine translation, text simplification, and paraphrasing tasks~\cite{ye2023dinoiser}. On the other hand, DINOISER would still struggle with discreteness in embedding spaces as it concentrates on manipulating noise scales whereas SEDD directly addresses this challenge through its loss function modelling. 

\section{Methodology}
% Describe the methods and algorithms used in your project. Explain why you chose these methods and how they are implemented. Provide details on any datasets, tools, or frameworks used.

% Are the methods and algorithms appropriate for the problem?
% Are the methods clearly explained and justified?
% Is the implementation described in sufficient detail, including datasets, tools, and frameworks, to ensure reproducibility?
\begin{figure*}
    \centering
    \includegraphics[width=\linewidth]{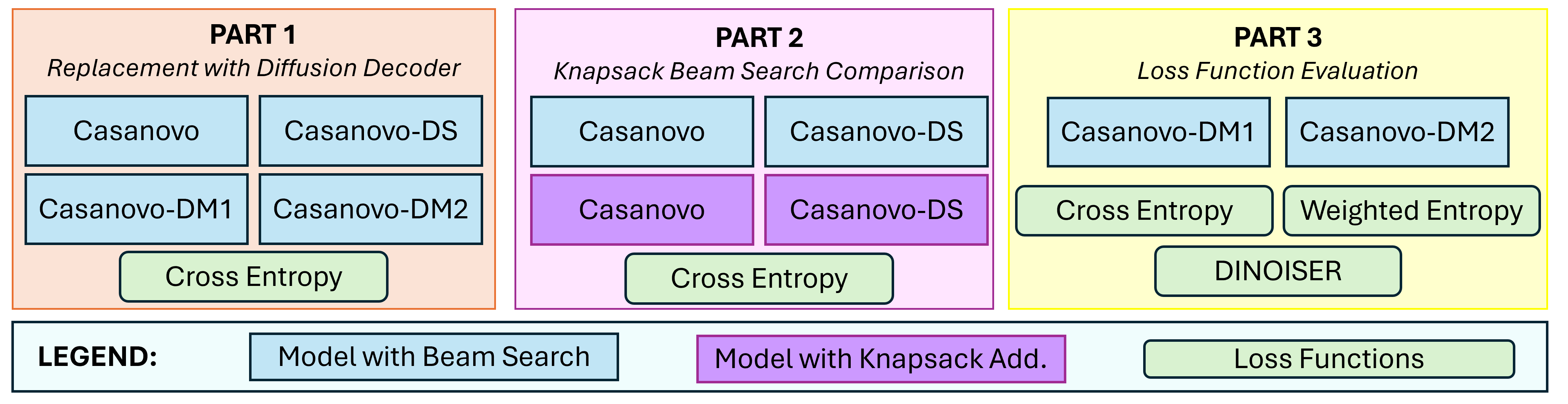}
    \caption{Overview of our methodology, with Part 1 exploring the replacement with different diffusion decoders in Casanovo, Part 2 trying knapsack with beam search, and Part 3 evaluating two of the diffusion model designs with different loss functions.}
  \label{fig:process-map-overview}
\end{figure*}

\subsection{Tools and Frameworks}
As seen in Figure~\ref{fig:process-map-overview}, we used Casanovo~\cite{casanovo-yilmaz2022novo} as our development framework as it has relatively lower computational demands but fairly high performance for the goal of peptide de novo sequencing compared to other state-of-the-art techniques like GraphNovo~\cite{graphnovo-mao2023mitigating} and InstaNovo~\cite{instanovo-eloff2025instanovo}. We replaced the autoregressive decoder with diffusion decoders and assessed discrete diffusion-specific losses inspired from score entropy~\cite{lou2023discrete} and DINOISER~\cite{ye2023dinoiser}. Although SEDD models have shown better performance compared to DINOISER, they are computationally more expensive and more difficult to work with. However, DINOISER is computationally less expensive and still has relatively high performance for discrete diffusion modelling. For training and evaluation, we used two NVIDIA GeForce RTX 4090 GPUs with 25 GB of memory each. Notably, GitHub Copilot~\cite{githubcopilot} was employed to help with the diffusion decoder code design, write code comments, and assist in debugging the code. The complete code is also attached as a zip folder with this report for reproducibility. 

\subsection{Dataset}
\label{sec:data}
For training and evaluation, we leveraged the same combined tryptic and non-tryptic dataset used in the latest version of Casanovo (Casanovo v4.2)~\cite{melendez2024accounting}. Tryptic refers to the use of the standard trypsin digestive enzyme for MS/MS~\cite{melendez2024accounting}. While trypsin is the standard digestive enzyme, other alternative enzymes have demonstrated improved peptide detection, making the combined dataset more representative for general peptide de novo sequencing~\cite{melendez2024accounting}. We obtained the data as annotated MGF files (a format for MS/MS data that is human readable~\cite{mgfdoc}) from the public Zenodo link on Casanovo's website~\cite{zenododata}. The combined dataset was created by obtaining data from MassIVE-KB v.2018-06-15 (tryptic) and MassIVE-KB v2.0.15 (non-tryptic)~\cite{wang2018assembling}. We employ the same respective train-validation-test split of 277,045, 27,437, and 200,000 PSMs~\cite{melendez2024accounting} for all of our experiments.

\subsection{Part 1: Replacement with Diffusion Decoder}
\begin{figure}[!h]
    \centering
    \includegraphics[width=\linewidth]{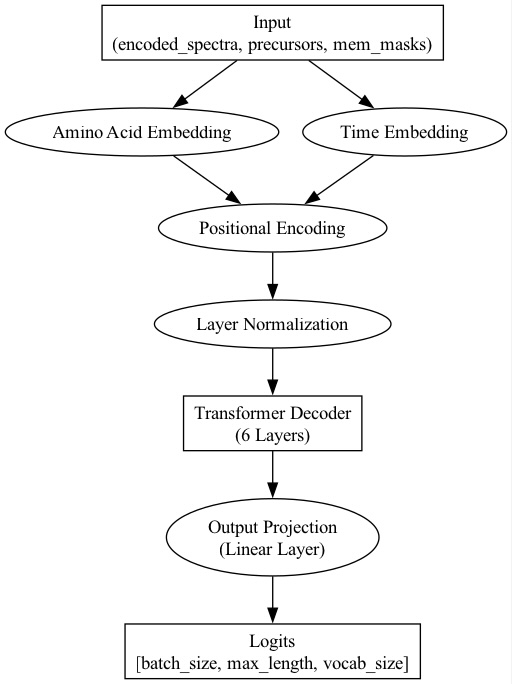}
    \caption{Casanovo-DM2.}
  \label{fig:diffusion-decoders-dm2}
\end{figure}

In the first part, we analyzed the merit of replacing the traditional autoregressive decoder with diffusion decoders. We utilized the original Casanovo with the transformer decoder (with 28.4 million parameters in the decoder) as the baseline for comparison. Three different diffusion decoder designs were integrated into the Casanovo framework and evaluated: Casanovo-DS (with 9.7 million parameters in the decoder), Casanovo-DM1 (with 29.5 million parameters in the decoder), and Casanovo-DM2 (with 29.6 million parameters in the decoder). Their architectures are shown below in Figure~\ref{fig:diffusion-decoders-ds}, Figure~\ref{fig:diffusion-decoders-dm1}, and Figure~\ref{fig:diffusion-decoders-dm2} respectively. We used the cross-entropy loss function for training along with the same training settings from Casanovo. 

\begin{figure}[!h]
    \centering
    \includegraphics[width=\linewidth]{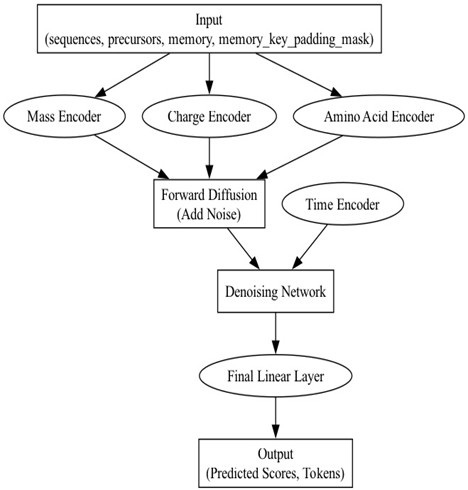}
    \caption{Casanovo-DS.}
  \label{fig:diffusion-decoders-ds}
\end{figure}

\begin{figure}[!h]
    \centering
    \includegraphics[width=0.7\linewidth]{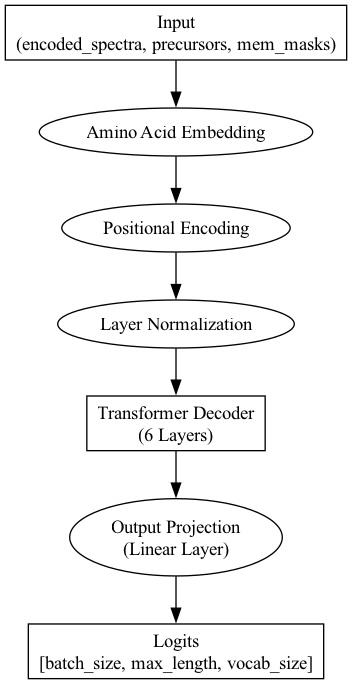}
    \caption{Casanovo-DM1.}
  \label{fig:diffusion-decoders-dm1}
\end{figure}

\subsection{Part 2: Knapsack Beam Search Comparison}
For this part, we assessed the benefit of using the knapsack method with the beam search algorithm. Inspired by its use in InstaNovo~\cite{instanovo-eloff2025instanovo}, we also incorporated it directly into the decoding process alongside beam search. The knapsack technique was used to iteratively build candidate sequences while considering constraints like precursor mass tolerance. Constraints were updated and reapplied after new candidates were generated and candidates were iteratively filtered if they exceeded the mass constraint. Noticeably, knapsack beam search was time demanding~\cite{instanovo-eloff2025instanovo} and as such, we only compared incorporating knapsack beam search for Casanovo-DS as it is smaller than Casanovo-DM1 and Casanovo-DM2. As a point of comparison, we also evaluated how the model's performance for Casanovo changed when knapsack beam search was implemented as well. Using the performance recorded in Part 1 with beam search, we calculated the shift in performance for both models. All model hyperparameters remained consistent between Part 1 and Part 2 to ensure a fair comparison.  

\subsection{Part 3: Loss Function Evaluation}
In this section, we examined incorporating different loss functions for our two most promising diffusion decoder designs, Casanovo-DM1 and Casanovo-DM2. Inspired by score entropy, we considered a weighted entropy loss function that combined the standard cross-entropy loss with an average entropy term that penalized the model if it is too uncertain, promoting more confident predictions. We also experimented with a discrete diffusion-specific loss from DINOISER that adds noise to the predicted logits. We compared the performance of using a weighted entropy loss function and the DINOISER loss function to values obtained from Part 1 with the cross-entropy loss function. Similar to Part 2, we kept the same model hyperparameters as before to ensure consistency in the comparison.

\section{Experiments and Results}
% Present the experiments conducted, including data preparation, parameter settings, and evaluation metrics. Display your results with tables, graphs, or figures as appropriate. Provide a clear interpretation of the results and discuss their significance.

% Are the experiments well-designed to evaluate the proposed methods or ideas?
% Are the experimental settings, datasets, and evaluation metrics clearly described?
% Are the results presented clearly with proper use of figures, tables, and graphs?

We compared models using peptide precision, peptide coverage, amino acid precision, and amino acid recall as these are the four most common evaluation metrics for peptide de novo sequencing and they provide a comprehensive view of the model performance for peptide sequencing at both the peptide and amino acid levels~\cite{deepnovo-tran2017novo,casanovo-yilmaz2022novo,graphnovo-mao2023mitigating,instanovo-eloff2025instanovo}. Peptide precision measures the fraction of predicted peptides that exactly match the true sequences over the total number of predicted spectra. Peptide coverage denotes the number of predicted spectra over the total number of available spectra. Amino acid precision assesses how many of the predicted amino acids are correct, calculated as the ratio of matched amino acids to total predicted amino acids. Lastly, amino acid recall evaluates the model’s sensitivity by measuring the proportion of true amino acids that are correctly predicted, calculated as the ratio of matched amino acids to the total number of actual amino acids in the true sequences. In all experiments, we used the same train-validation-test split described in Section~\ref{sec:data} along with the default Casanovo model hyperparameters. To compute statistical significance for the best model improvement, we used scipy.stats wilcoxon~\cite{scipy} for the Wilcoxon signed-rank test~\cite{rainio2024evaluation}. 

\subsection{Part 1}
\begin{figure*}
    \centering
    \includegraphics[width=\linewidth]{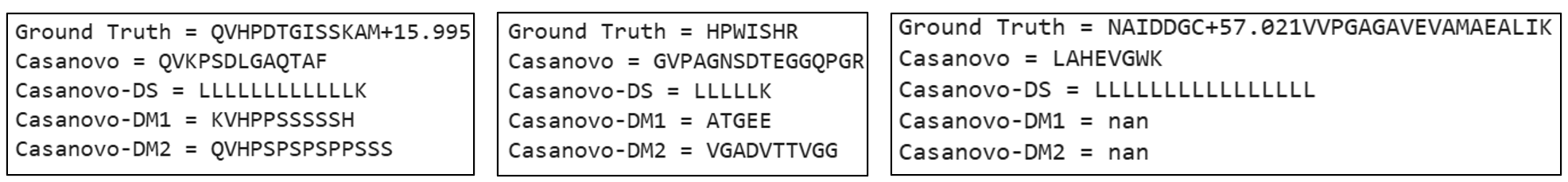}
    \caption{Comparison of sample predicted sequences from models in Part 1 highlighting the impact of using three different decoder designs (Casanovo-DS, Casanovo-DM1, and Casanovo-DM2) versus an autoregressive transformer decoder (Casanovo).}
  \label{fig:part1-comparison}
\end{figure*}

In this part, we evaluated the performance of Casanovo (with a transformer decoder) along with three other variants of Casanovo that used different decoder designs (seen in Figure~\ref{fig:diffusion-decoders-ds}, Figure~\ref{fig:diffusion-decoders-dm1}, and Figure~\ref{fig:diffusion-decoders-dm2}). As seen in Table~\ref{tab:part1-results}, simply replacing the transformer decoder with different decoder designs led to worse performance across all four metrics. A qualitative examination of the outputs showed interesting findings as Casanovo-DM1 and Casanovo-DM2 generally produced more accurate sequences than Casanovo-DS (see left of Figure~\ref{fig:part1-comparison}) but tended to fail more than Casanovo-DS (see right part of Figure~\ref{fig:part1-comparison}) which led to an overall lower quantitative metric score. Interestingly, all three diffusion models sometimes produced sequences of more similar size to the ground truth than Casanovo in failure cases (see middle part of Figure~\ref{fig:part1-comparison}).

\begin{table*}
\centering
\caption{Model performance comparison for replacing the transformer decoder (in Casanovo) with three different diffusion decoder designs (in Casanovo-DS, Casanovo-DM1, Casanovo-DM2) where AA stands for Amino Acid.}
\begin{tabular}{ l c c c c } 
 \hline
 Model & Peptide Precision & Peptide Coverage & AA Precision & AA Recall \\ 
 \hline
 Casanovo & 0.037 & 0.690 & 0.080 & 0.081 \\ 
 Casanovo-DS & 0.000 & 0.000 & 0.042 & 0.041 \\ 
 Casanovo-DM1 & 0.000 & 0.000 & 0.031 & 0.018 \\ 
 Casanovo-DM2 & 0.000 & 0.000 & 0.037 & 0.027 \\ 
 \hline
\end{tabular}
\label{tab:part1-results}
\end{table*}

\subsection{Part 2}
\begin{figure*}
    \centering
    \includegraphics[width=\linewidth]{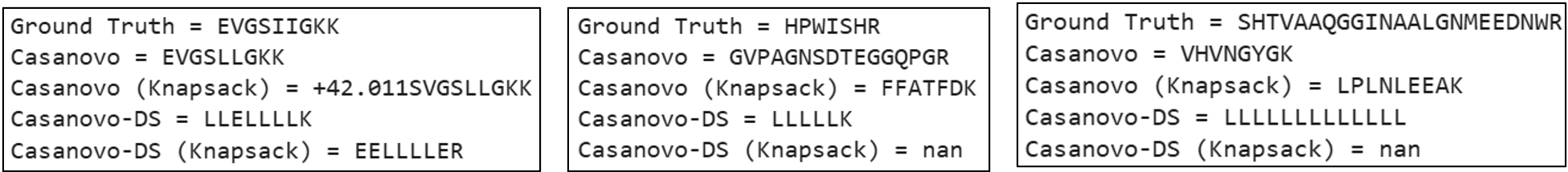}
    \caption{Comparison of sample predicted sequences from models in Part 2 showcasing the impact of knapsack beam search decoding.}
  \label{fig:part2-comparison}
\end{figure*}

In Part 2, we evaluated the merit of replacing beam search with knapsack beam search. Given the time cost of knapsack beam search, we experimented with only the Casanovo and Casanovo-DS models. The quantitative results in Table~\ref{tab:part2-results} demonstrate that using knapsack beam search resulted in worse metric performance across all four metrics for both Casanovo and Casanovo-DS. As seen in Figure~\ref{fig:part2-comparison}, knapsack beam search greatly changed the predicted sequence for these two models where Casanovo-DS with Knapsack generally produced qualitatively better sequences, but had more errors than Casanovo-DS (using only beam search). On the other hand, Casanovo with Knapsack generally led to worse performance but in failure cases, the model sometimes produced sequences with more similar size to the ground truth than Casanovo (with only beam search). In terms of time cost, using knapsack beam search increased the training time for both models from roughly 1.5 hours to 35 hours.

\begin{table*}
\centering
\caption{Model performance comparison after switching to knapsack beam search for Casanovo and Casanovo-DS with the change in performance from beam search indicated in brackets (negative value indicates that using beam search resulted in a higher metric value than using knapsack beam search), where AA refers to Amino Acid.}
\begin{tabular}{ l c c c c } 
 \hline
 Model & Peptide Precision & Peptide Coverage & AA Precision & AA Recall \\ 
 \hline
 Casanovo & 0.000 (-0.037) & 0.000 (-0.690) & 0.043 (-0.037) & 0.035 (-0.046) \\ 
 Casanovo-DS & 0.000 (0.000) & 0.000 (0.000) & 0.018 (-0.024) & 0.000 (-0.041) \\ 
 \hline
\end{tabular}
\label{tab:part2-results}
\end{table*}

\subsection{Part 3}
Part 3 evaluated different loss functions for training different diffusion decoder designs. Despite their lower performance in Part 1, we chose to experiment with Casanovo-DM1 and Casanovo-DM2 instead of Casanovo-DS as they had more promising sequencing results qualitatively. Unfortunately, the peptide precision and peptide coverage remained at 0 across all three loss functions. However, the amino acid precision and recall metrics were improved by using different loss functions as exhibited in Table~\ref{tab:part3-results}. Interestingly, the optimal loss function for both Casanovo-DM1 and Casanovo-DM2 was DINOISER with higher amino acid precision and recall metrics compared to using cross-entropy or weighted entropy loss functions. Although using weighted entropy was better than using cross-entropy for Casanovo-DM1, the performance for Casanovo-DM2 was similar for both loss functions. Overall, Casanovo-DM2 with the DINOISER loss function achieved the highest amino acid precision and recall. Although the amino acid precision for the best setup (0.070) was still lower than Casanovo's at 0.080, the amino acid recall with the best configuration (0.454) far surpassed that of Casanovo at 0.081 with a p-value $<$ 0.001, indicating statistical significance. 

\begin{figure*}
    \centering
    \includegraphics[width=\linewidth]{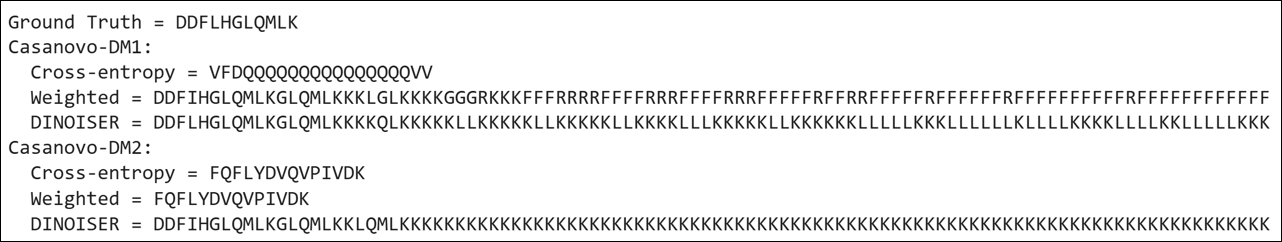}
    \caption{Sample predicted sequences produced by Casanovo-DM1 and Casanovo-DM2 using the three loss functions where Weighted refers to the weighted entropy loss function.}
  \label{fig:part3-sample}
\end{figure*}

A qualitative examination of the predicted sequences does raise concerns as most of the outputs are much longer compared to the ground truth sequence (example in Figure~\ref{fig:part3-sample}). Although the amino acid sequence more closely matches the ground truth at the start of the sequence, the remainder of the sequence appears to be noise. Despite the noise, the amino acid recall still remained high as this metric is based on the number of amino acids in the true sequence rather than the predicted sequence. Since the predicted sequence typically matches the true sequence for the true sequence's length, it leads to a higher recall.

\begin{table*}
\centering
\caption{Amino Acid (AA) performance of Casanovo-DM1 and Casanovo-DM2 for different loss functions with best metrics obtained by Casanovo-DM2 with DINOISER.}
\begin{tabular}{ l c c c c } 
 \hline
 \multirow{2}{*}{Loss Function} & \multicolumn{2}{c}{Casanovo-DM1} & \multicolumn{2}{c}{Casanovo-DM2} \\ 
  & AA Precision & AA Recall & AA Precision & AA Recall \\ 
 \hline
 Cross-Entropy & 0.031 & 0.018 & 0.037 & 0.027 \\ 
 Weighted Entropy & 0.066 & 0.423 & 0.037 & 0.026\\ 
 DINOISER & 0.070 & 0.451 & 0.070 & 0.454\\ 
 \hline
\end{tabular}
\label{tab:part3-results}
\end{table*}

\begin{table*}
\centering
\caption{Amino Acid (AA) performance of Casanovo-DM1 and Casanovo-DM2 for different loss functions with best metrics obtained by Casanovo-DM2 with DINOISER.}
\begin{tabular}{ l c c} 
 \hline
 Model & AA Precision & AA Recall \\ 
 \hline
 Casanovo (Cross-Entropy) & \textbf{0.080} & 0.081 \\
 Casanovo-DS (Cross-Entropy) & 0.042 & 0.041 \\
 Casanovo-DM1 (Cross-Entropy)& 0.031 & 0.018 \\
 Casanovo-DM2 (Cross-Entropy)& 0.037 & 0.027 \\
 Casanovo-DM1 (Weighted Entropy) & 0.066 & 0.0423 \\
 Casanovo-DM2 (Weighted Entropy) & 0.037 & 0.026 \\
 Casanovo-DM1 (DINOISER) & 0.070 & 0.451 \\
 Casanovo-DM2 (DINOISER) & 0.070 & \textbf{0.454} \\
 \hline
\end{tabular}
\label{tab:part3-results}
\end{table*}

\section{Discussion and Conclusion}
% Analyze the strengths and limitations of your approach. Compare your results to existing work or state-of-the-art methods. Discuss any unexpected findings or challenges encountered during the project. Summarize the key findings and contributions of your project. Suggest possible future directions for research based on your work.

% Does the project demonstrate a meaningful contribution to the field, such as new insights, methods, or applications?
% Does it discuss the potential impact, strengths, limitations, and future research directions?

This paper studies the problem of diffusion decoding in peptide de novo sequencing. We investigated using different diffusion decoders as an alternative to conventional autoregressive decoders, compared the merit of using knapsack beam search against beam search, and assessed the performance of various loss functions relative to the standard cross-entropy loss. In our work, we showed that simply replacing the transformer decoder with a diffusion decoder was inadequate for obtaining high performance and switching from beam search to knapsack beam search incured a high training time cost (from 1.5 hours to 35 hours) with worse performance. However, by incorporating an appropriate loss function for discrete diffusion (DINOISER), we were able to increase amino acid recall by 0.373 with statistical significance compared to Casanovo (with an autoregressive transformer decoder). 

The main advantage of our approach is the ability to decode from any part of the peptide sequence whereas current state-of-the-art methods can only decode sequentially from one end of the sequence. Additionally, our method is less computationally demanding and can be trained in a single step, unlike the two-step resource-intensive training process used in InstaNovo, which constrains the performance of the diffusion model to rely on the training of a separate transformer-based model. However, our approach does have limitations, including much lower peptide precision and coverage. Our work is also limited by difficulties in acquiring the necessary GPUs to train other published methods, such as GraphNovo and InstaNovo, making it challenging to fairly compare our approach to theirs. Running their pretrained models on our test set is not a fair comparison, as their models were trained on larger and different datasets and previous studies have demonstrated that models trained on higher-quality data can yield better results, not necessarily due to superior architecture, but because of the quality of the data used during training~\cite{deepnovo-tran2017novo}. Another limitation is that our final best model, obtained using Casanovo-DM2 with a DINOISER loss function, predicted sequences that were much longer than the true sequence and contained noise at the end, likely exceeding the desired mass capacity. 

Future work could address this challenge by finding better ways to truncate the leftover noise through potentially leveraging knapsack filtering when developing the sequence. Although our work showed decreased performance when using the knapsack algorithm during beam search, there may be better ways to incorporate this algorithm to address this challenge. Furthermore, we used the initial Casanovo hyperparameters but hyperparameter tuning should be explored for improved performance for these diffusion decoder-based models. For generalization, future work should also include training and inferencing with other data sources such as those in GraphNovo~\cite{graphnovo-mao2023mitigating} and InstaNovo~\cite{instanovo-eloff2025instanovo}. Even though it may be computationally expensive to train GraphNovo or InstaNovo, developing and benchmarking models using the same data would allow for a fair comparison of model performance. 

\bibliography{references}
\bibliographystyle{icml2025}

%%%%%%%%%%%%%%%%%%%%%%%%%%%%%%%%%%%%%%%%%%%%%%%%%%%%%%%%%%%%%%%%%%%%%%%%%%%%%%%
%%%%%%%%%%%%%%%%%%%%%%%%%%%%%%%%%%%%%%%%%%%%%%%%%%%%%%%%%%%%%%%%%%%%%%%%%%%%%%%
% APPENDIX
%%%%%%%%%%%%%%%%%%%%%%%%%%%%%%%%%%%%%%%%%%%%%%%%%%%%%%%%%%%%%%%%%%%%%%%%%%%%%%%
%%%%%%%%%%%%%%%%%%%%%%%%%%%%%%%%%%%%%%%%%%%%%%%%%%%%%%%%%%%%%%%%%%%%%%%%%%%%%%%
\newpage

\end{document}